%% file: paper.tex
\documentclass[10pt,twocolumn,letterpaper]{article}

\usepackage[pagenumbers]{cvpr} 

\usepackage{graphicx}
\usepackage{amsmath}
\usepackage{amssymb}
\usepackage{booktabs}

\usepackage{float}
\usepackage{textcomp, gensymb}
\usepackage{multirow}
\usepackage{makecell}

\usepackage[pagebackref,breaklinks,colorlinks]{hyperref}

\usepackage{algorithm}
\usepackage[noend]{algpseudocode}

\usepackage{array}
\newcolumntype{P}[1]{>{\centering\arraybackslash}p{#1}}

\makeatletter
\newcommand{\linebreakand}{%
  \end{@IEEEauthorhalign}
  \hfill\mbox{}\par
  \mbox{}\hfill\begin{@IEEEauthorhalign}
}
\makeatother

\usepackage{adjustbox}
\newcommand\animage{\adjustbox{valign=m,vspace=0.5pt}{\includegraphics[height=65pt]{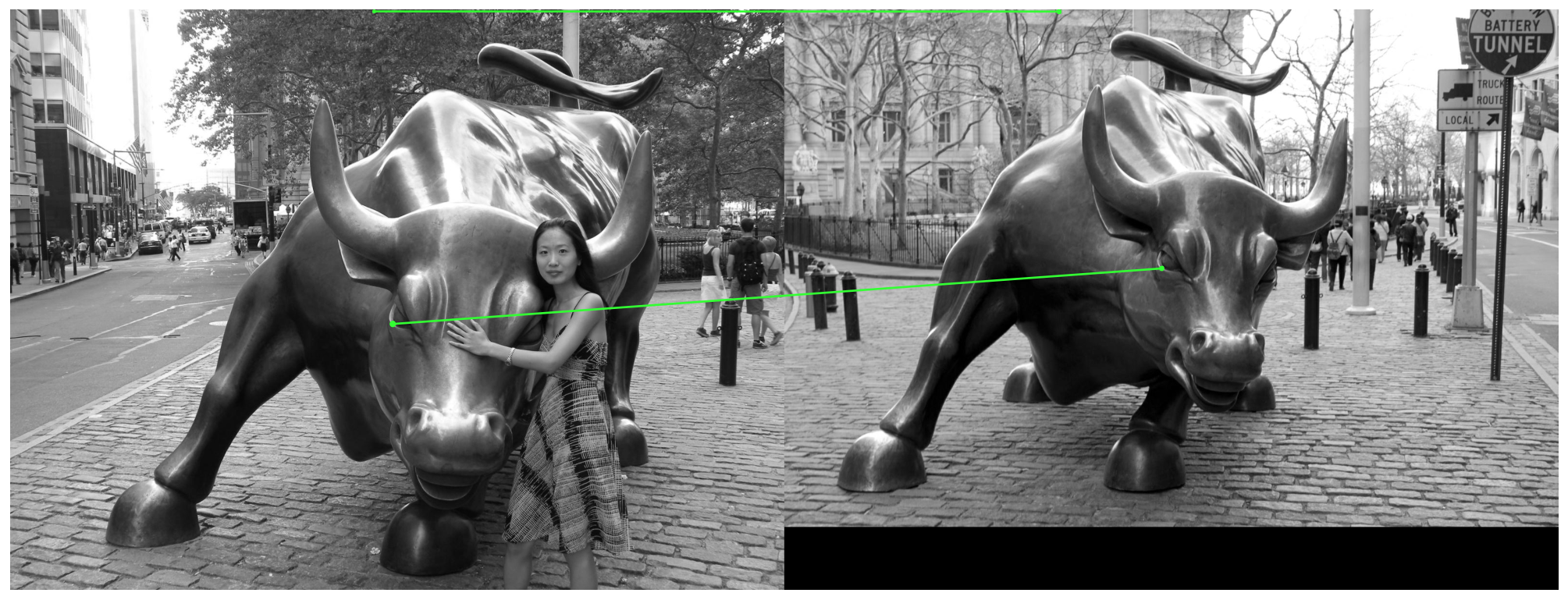}}}

\newcommand\animageone{\adjustbox{valign=m,vspace=0.5pt}{\includegraphics[height=65pt]{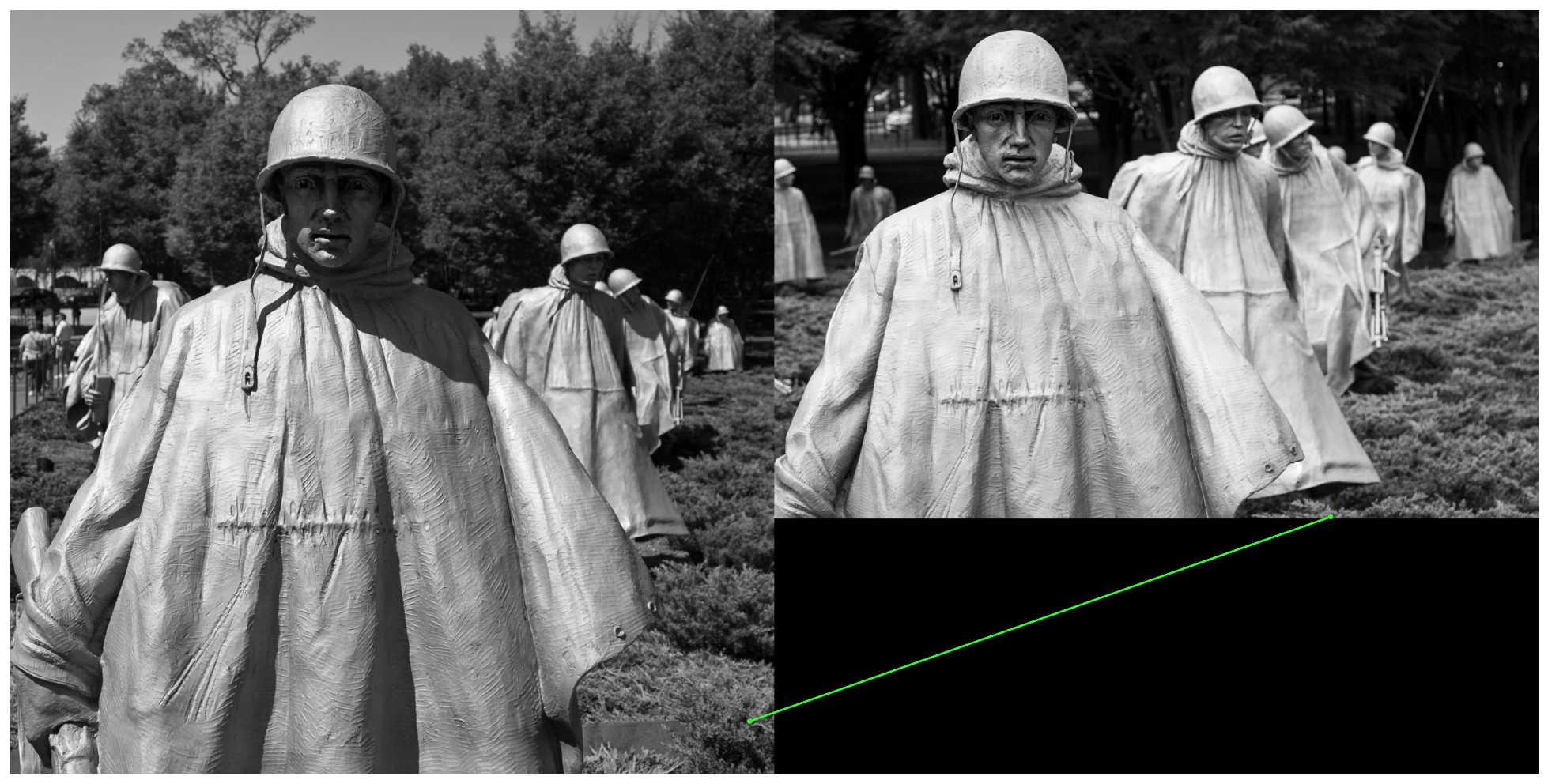}}}

\newcommand\animagetwo{\adjustbox{valign=m,vspace=0.5pt}{\includegraphics[height=65pt]{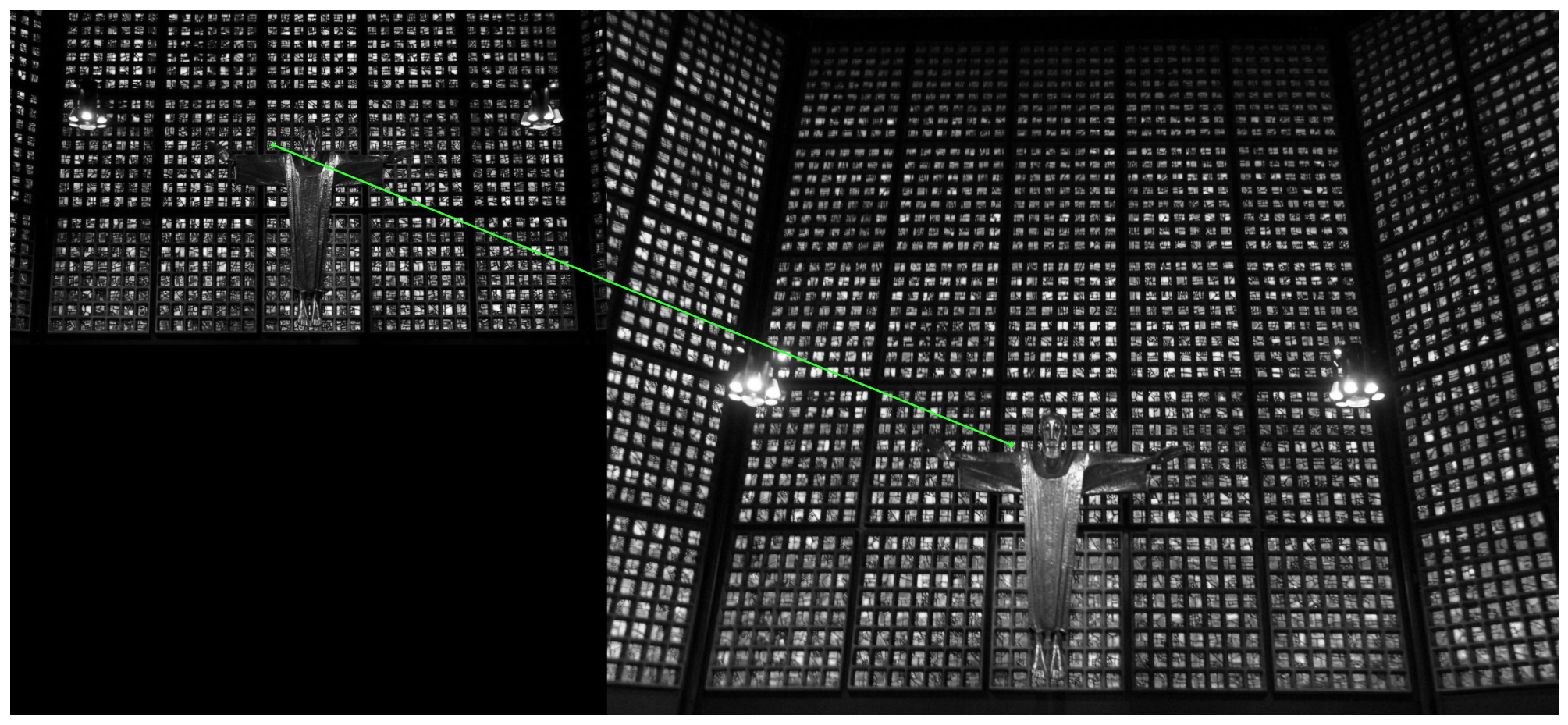}}}

\newcommand\animagegeo{\adjustbox{valign=m,vspace=0.5pt}{\includegraphics[height=65pt]{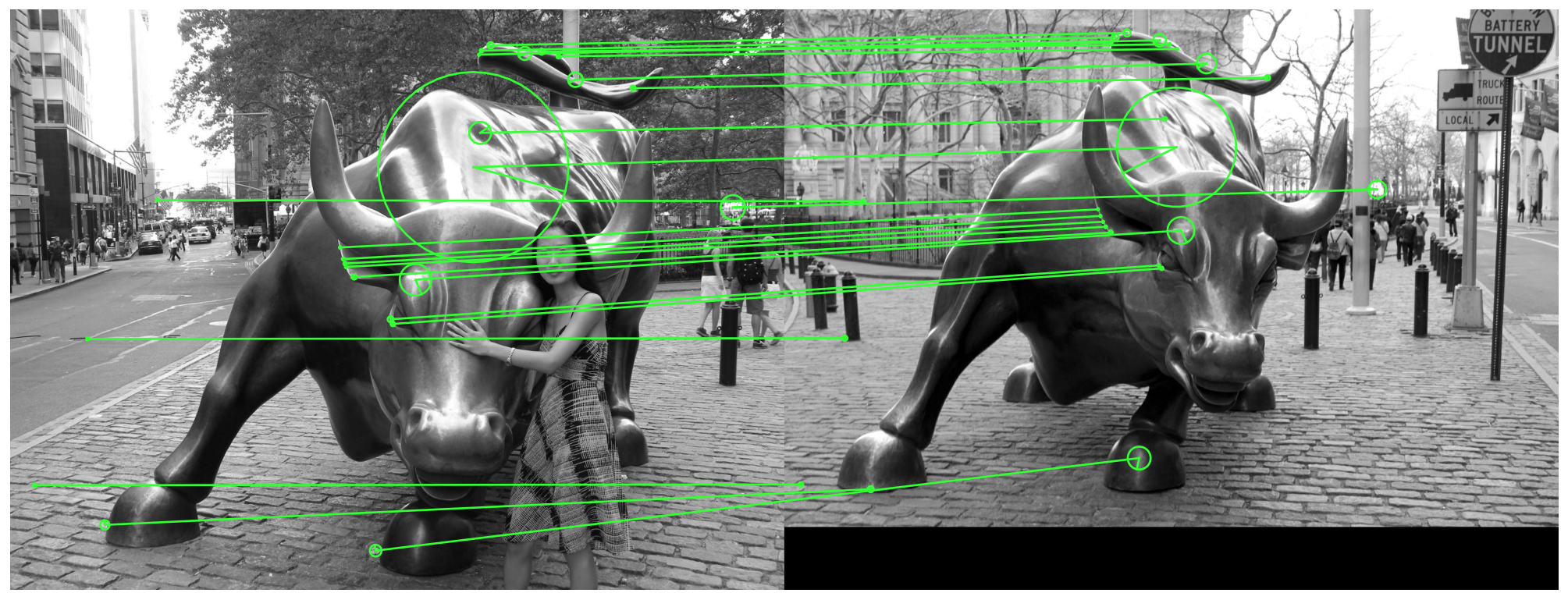}}}

\newcommand\animageonegeo{\adjustbox{valign=m,vspace=0.5pt}{\includegraphics[height=65pt]{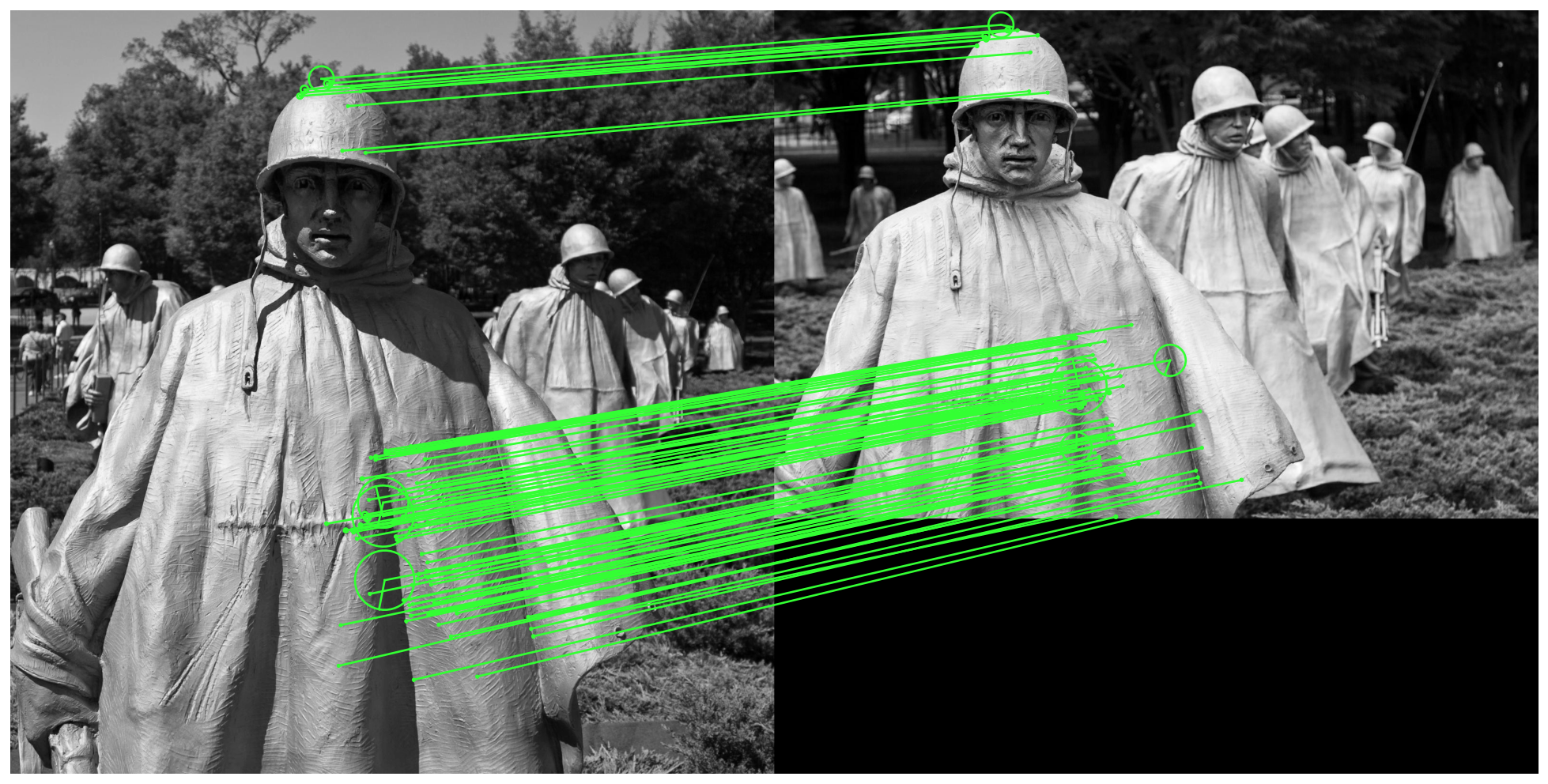}}}

\newcommand\animagetwogeo{\adjustbox{valign=m,vspace=0.5pt}{\includegraphics[height=65pt]{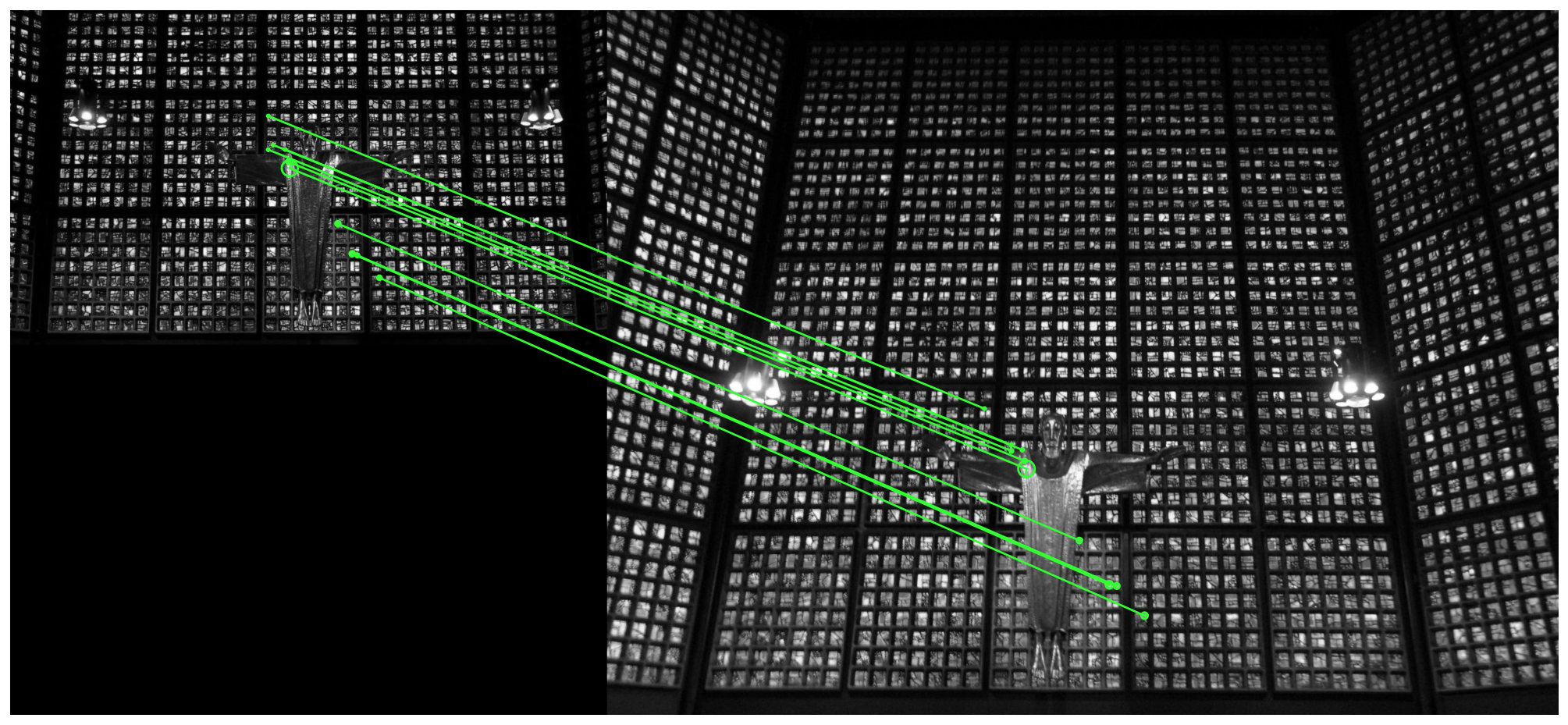}}}

\usepackage[capitalize]{cleveref}
\crefname{section}{Sec.}{Secs.}
\Crefname{section}{Section}{Sections}
\Crefname{table}{Table}{Tables}
\crefname{table}{Tab.}{Tabs.}

\def\cvprPaperID{*****} 
\def\confName{CVPR}
\def\confYear{2022}

\begin{document}

\title{OpenGlue: Open Source Graph Neural Net Based Pipeline for Image Matching}

\author{
    Ostap Viniavskyi\\
    The Machine Learning Lab\\ Ukrainian Catholic University\\ Lviv, Ukraine\\
    {\tt\small viniavskyi@ucu.edu.ua}
    \and
    Mariia Dobko\\
    The Machine Learning Lab\\ Ukrainian Catholic University\\ Lviv, Ukraine\\
    {\tt\small dobko\_m@ucu.edu.ua}
    \and
    Dmytro Mishkin\\
    Faculty of Electrical Engineering\\
    Czech Technical University\\
    Prague, Czech Republic\\
    {\tt\small  mishkdmy@fel.cvut.cz}
    \and 
    Oles Dobosevych\\
    The Machine Learning Lab\\ Ukrainian Catholic University\\ Lviv, Ukraine\\
    {\tt\small  dobosevych@ucu.edu.ua}
}

\maketitle

\begin{abstract}
We present OpenGlue: a free open-source framework for image matching, that uses a Graph Neural Network-based matcher inspired by SuperGlue~\cite{sarlin20superglue}. We show that including additional geometrical information, such as local feature scale, orientation, and affine geometry, when available (e.g. for SIFT features), significantly improves the performance of the OpenGlue matcher. We study the influence of the various attention mechanisms on accuracy and speed. We also present a simple architectural improvement by combining local descriptors with context-aware descriptors. The code and pretrained OpenGlue models for the different local features are publicly available \footnote{\url{https://github.com/ucuapps/OpenGlue}}.  
\end{abstract}


\input{sections/Introduction}
\input{sections/Method}
\input{sections/Datasets}
\input{sections/Experiments}
\input{sections/Conclusions}

\section*{Acknowledgements} 
Authors thank Ukrainian Catholic University for providing necessary computing resources. D.\ Mishkin is supported by OP VVV funded project CZ.02.1.01/0.0/0.0/$16\_019$/0000765 ``Research Center for Informatics''. We also express gratitude to James Pritts for his help and support.

{\small
\bibliographystyle{ieee_fullname}
\bibliography{literature}
}

\end{document}

%% file: sections/Introduction.tex
\section{Introduction}
Image matching is a fundamental problem in multiple-view geometry \cite{Zitovaimageregistration} which attempts to overlay two or more images of one scene or object. Detecting correspondences between images is a primary step towards solving many problems in computer vision: scene reconstruction~\cite{RomeInDay2009,Heinly15,COLMAP2016}, visual localization~\cite{ActiveSearch2012,Sattler18,Lynen19}, SLAM~\cite{Mur15,DeTone17a,SuperPoint2017}, image retrieval~\cite{DBLP:journals/ijcv/Lowe04,NetVLAD2016,DELF2017}. The most widely adopted pipeline for image matching~\cite{Pritchett1998, mishkin2021learning} consists of four consecutive steps: feature detection, feature description, feature matching and filtering, geometric transformation estimation. 
\par
The SuperGlue \cite{sarlin20superglue} approach to image matching fits into the standard image matching pipeline replacing classical Nearest Neighbor-based methods. It tackles keypoints matching by formulating it as an end-to-end classification task solved by an Attention-based \cite{DBLP:conf/nips/VaswaniSPUJGKP17} Graph Neural Network (AGNN).
\par
We propose OpenGlue, an open-source framework with an AGNN at the core based on SuperGlue image matcher. Our solution offers several alterations to the original architecture of SuperGlue, whose effect we study in this paper. In contrast to the original method, OpenGlue makes use of an additional geometric information, which can be extracted from many local feature detectors and descriptors, e.g. SIFT~\cite{DBLP:journals/ijcv/Lowe04} -- like orientation, scale and so on. In addition, we study the effect of using Non-Maximum Suppression (NMS) over detected keypoints when applying OpenGlue matcher. We show that applying NMS both at training and testing improve the results of matching and downstream tasks. 
\par
Also, we study various efficient attention mechanisms that help speed up the training and inference while also saving memory immensely. We mainly focus on linear~\cite{katharopoulos-et-al-2020} and FAVOR+~\cite{choromanski2020rethinking} versions of attention and show how they contribute to the faster computation time and memory usage.

\begin{figure*}[!t]
\centering
\includegraphics[width=6.5in]{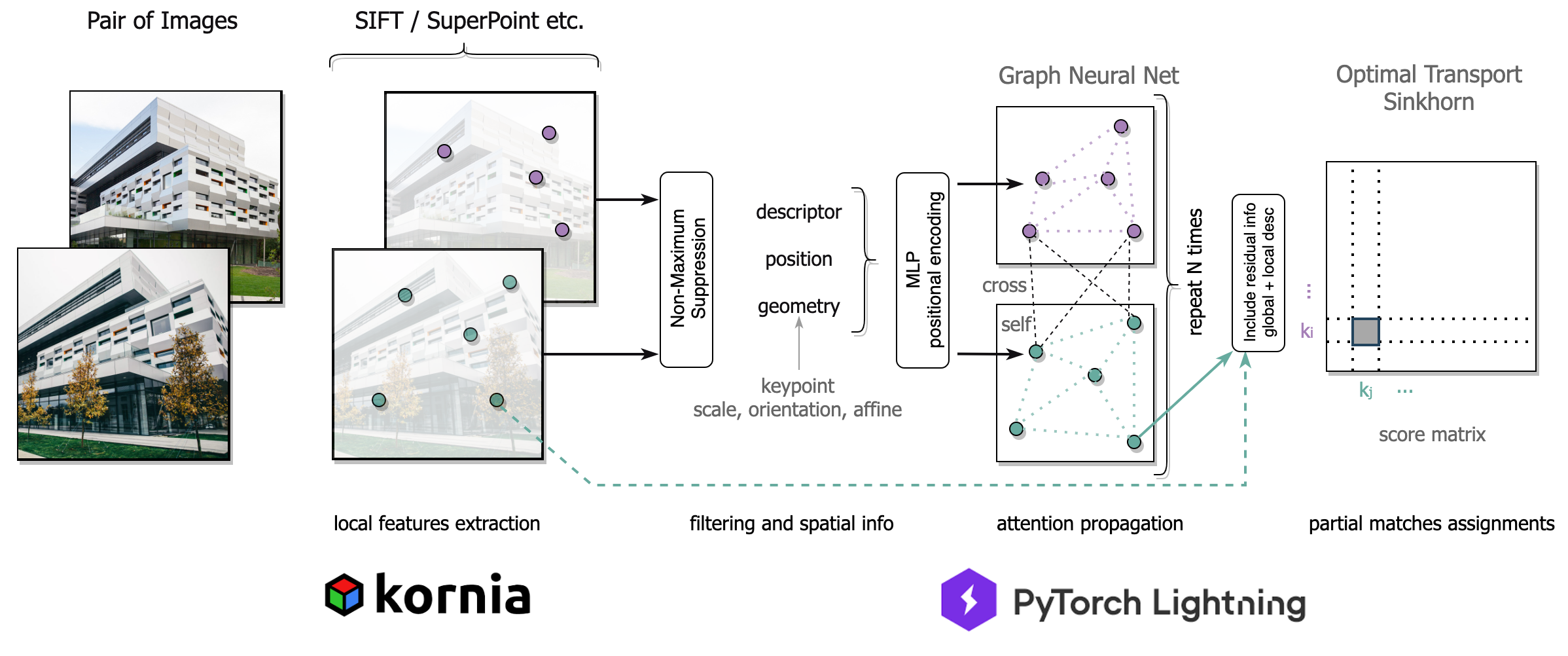}
\vspace{-1.5em}
\caption{\textbf{OpenGlue Pipeline} The main stages include: local features detection and description, filtering keypoints with non-maximum suppression, encoding of positional and geometric information, Attentional Graph Neural Network (AGNN) for propagation of contextual information among keypoints in both images, Optimal Transport method for matches assignment. We introduce an open-source version based on Kornia library and PyTorch Lightning framework.}
\label{fig:pipeline}
\end{figure*}

\par 
Our contribution can be summed up as follows:
\begin{itemize}
  \item Presented an open-source framework with flexible components that can be adapted to various needs. The implemented pipeline supports a four-stage approach to image matching and provides an opportunity to substitute or customize each step. For example, at the first stage of local feature extraction user can choose one of the several extractors or add a custom one following a standard showcased interface. We include classical local feature extractors together with learnable methods and provide options for customizing their architecture.
  \item Improved SuperGlue results for SIFT-extracted features by adding non-maximum suppression over detected keypoints.
  \item Showed the increase in matcher performance by adding keypoint geometrical information into positional encoding.
  \item Tested the impact of different attention mechanisms.
  \item Combined local and global descriptors in the final stage before matching to include residual information and improve the matching results.
\end{itemize}

%% file: sections/Method.tex
\section{Method}


The general pipeline of OpenGlue is shown in Figure \ref{fig:pipeline}. Here, we present each component in detail describing our added components over SuperGlue.

\subsection{Local features extraction}

There are several ways to approach local features extraction. For example, using handcrafted methods for keypoints detection and description. However, some of them perform only detection task: Harris corner detector \cite{Harris88acombined},  GFTT \cite{JianboShi1994}, FAST \cite{Rosten06machinelearning}, MSER \cite{Matas02robustwide}; some can only provide description: BRIEF \cite{Calonder10brief:binary}, while others can do both: ORB \cite{6126544}, SIFT \cite{DBLP:conf/iccv/Lowe99}, SURF \cite{DBLP:conf/eccv/BayTG06}.
Feature detection and description steps benefit a lot from applying deep learning. Features produced by such methods as SuperPoint \cite{DBLP:conf/cvpr/DeToneMR18}, D2-Net \cite{DBLP:conf/cvpr/DusmanuRPPSTS19}, HardNet \cite{DBLP:conf/nips/MishchukMRM17} set new state-of-the-art results in various image matching benchmarks. 
\par
We use three different methods for detection and description of keypoints in our implementation to study the performance of OpenGlue in various settings: SIFT (hand-crafted method that bundles a feature detector and descriptor), Difference-of-Gaussians (DoG) detection with AffNet \cite{AffNet2018} for affine shape estimation and HardNet \cite{HardNet2017} descriptor (called for short DoG-AffNet-HardNet, hand-crafted detector with learned local affine frame estimation and descriptor), SuperPoint (end-to-end deep network for simultaneous detection and description). 
This variety of local feature extraction methods provides flexibility for a user to select the best approach for their task. While learnable methods have shown outstanding performance in different multiple-view problems, classical approaches are still reasonable competitors, often winning in speed and resource expenses.  

\par
SuperPoint performs detection and description with a single CNN model that shares a backbone and has separate heads for predicting keypoints and describing them. The detector head is trained in a self-supervised manner by using a process of Homographic Adaptation, where ground-truth keypoints are generated by making detections on different warped versions of the same image and combining them together. The model learns to predict keypoints locations by using cross-entropy loss which is used for pixel-wise classification. The descriptor head is trained by using metric learning approach. By applying random homographies, one can establish ground-truth correspondences between points in an image and its warped version. The description head objective function tries to minimize the distance between descriptors of such two keypoints. The distance to descriptors of all others keypoints is maximized.
\par
There are a few versions of open-source SuperPoint weights, released by You-Yi Jau and Rui Zhu \cite{SuperpointWeights}. Some were produced on KITTI \cite{Geiger2013IJRR} while others on COCO \cite{lin2014microsoft} dataset. In contrast to the SuperPoint from original SuperGlue approach, the architecture of the model includes batch normalization in each block. We do not train the SuperPoint from scratch, instead we use the open-source weights for our experiments.

\subsection{Geometry Information in Positional Encoding}
SuperGlue proposed to use a Multilayer Perceptron (MLP) to encode the positional information about keypoints. This encoding can be combined with visual information (descriptors) and propagated for joint training. In OpenGlue, a feed-forward network generates a positional encoding for each keypoint in the image. The number of layers is configurable, while a default depth is three layers, where dense layer is followed by RELU activation and batch normalization. 
\par
We enable the use of the estimated geometry from keypoints detectors in the positional encoding. René Ranftl and Vladlen Koltun previously explored the usage of keypoint geometry as an input for neural network that predicts weights for learnable IRLS\cite{Ranftl2018}. Similarly, we exploit the ability of some detectors to capture the orientation and the spatial relationships of the features to improve the OpenGlue matcher. We provide scale, rotation, the combination of both, or local affine geometry of the keypoint as an additional input to the MLP for positional encoding, depending on the information available from feature extractor. The orientation of keypoint is encoded as cosine and sine of the orientation angle, while the scale is transformed with logarithmic function. Local affine geometry of a keypoint is represented by an affine matrix which we first normalize such that its determinant is 1, and provide separately flattened transformation matrix and log-scale.

\subsection{Non-maximum Suppression}
When training OpenGlue, we select only a fixed number of keypoints in the images to perform efficient batching of training examples. At inference time, the number of keypoint is also limited due to available resources constraints. Keypoints are selected based on their response score from the detection algorithm. One of the problems with local features from such methods as DoG is that keypoints with highest responses are often gathered in the small part of the image. When combined with top-response keypoints filtering, this leaves a large areas in the images with little to no detected keypoints. We show significant improvements in matching when using Non-Maximum Suppression on SIFT and DoG-AffNet-HardNet keypoints both at training and testing time. The same procedure is applied for SuperPoint. NMS filters the keypoint candidates from the detection stage and keeps only those with the largest response in their neighborhood. The kernel size for suppression is chosen as 9.  

\subsection{Attention Graph Neural Network}
SuperGlue uses an undirected graph over all the keypoints in both images. The edges are split into two separate sets, one connecting keypoints that lies in a single image, while the other edges connect keypoints across images. Attention Graph Neural Network propagates information only from keypoints that are connected by one type of edges at a single layer. At odd layers, information is propagated using self-edges, while at even layers from cross-edges. AGNN propagates information from one node to the other by using message passing \cite{DBLP:conf/icml/GilmerSRVD17}. The message is computed by using dot-product attention, where the query is an intermediate representation of keypoint, and keys and values are computed based on intermediate representations of message sources. The original representation of each keypoint is composed of visual descriptor along with positional information embedded in a higher-dimensional space with Multi-Layer Perceptron. The representation after the last layer is linearly projected to obtain the final global context-aware descriptor for each keypoint. Descriptors are kept unnormalized. The norm reflects the matching confidence for each keypoint.
\par
\textbf{Linear attention:} In an attempt to make OpenGlue faster and less demanding for GPU memory we experiment with different efficient attention types. In particular, we use linear attention \cite{katharopoulos-et-al-2020}, which allows decomposing the attention computations in such a way, that there is no need to store the complete attention weights matrix. We use $ x' = elu(x) + 1$ transformation for a kernel, that computes similarity between each query and key in an attention procedure.

\par
\textbf{FAVOR+:} Additionally, we exploit a recent progress in efficient attention types \cite{choromanski2020rethinking}, by trying to approximate softmax kernel with random feature maps. FAVOR+ uses positive orthogonal random features to transform each query and key separately, such that the dot product of their transformations approximates softmax kernel. In this way, we can use the benefits of linear attention, while approximating the original softmax attention.

\subsection{Residual Information}

We present an additional learnable vector $\alpha$ which is used for combining the global contextual descriptor with the local one. The final keypoint descriptor is a convex combination of both global and local descriptors, where each dimension is weighted separately according to $\alpha_i$.  The descriptors are then updated in a following manner:
\[ gdesc = \sigma(\alpha) * gdesc + (1.0 - \sigma(\alpha)) * ldesc\]
where $ldesc$ represents local descriptor, an input to AGNN, $gdesc$ represents a context-aware  descriptor, an output of AGNN, and $\sigma$ is a sigmoid function.

\subsection{Optimal Transport}
Optimal Matching Layer computes an assignment matrix representing the matching probabilities for each possible correspondence. Each keypoint in one image can be matched to only one keypoint in the other image or have no match at all because of viewpoint change and occlusion. To account for keypoints that cannot be matched, SuperGlue adds an auxiliary dustbin keypoint on each image that will match any unmatchable keypoints in other images. Matching score between any keypoint and dustbin, as well as between two dustbin is a learnable parameter. The partial assignment matrix is computed using a differentiable Sinkhorn algorithm \cite{DBLP:conf/nips/Cuturi13}.

\section{Implementation Details}
We develop an open source version of SuperGlue, the OpenGlue, which can be used in academic or commercial applications. It is based on PyTorch Lightning \cite{FalconPyTorchLightning2019} framework and enables user to train, predict and evaluate the model. For local feature extraction, our interface supports Kornia \cite{eriba2019kornia} detectors and descriptors along with our version of SuperPoint. In this section we describe the technical details of our implementation.

\subsection{Structure and Interface}

Our motivation was to develop a library that allows users to fully use the SuperGlue model with ability to add support for custom components. The pipeline is extremely configurable, however, we provide some initial default configurations to simplify the use for common purposes. We follow the generally established design goals of keeping interfaces simple and consistent to give a chance for implementation of additional features and adaptation to new scenarios.
\par
\begin{figure}[!t]
\centering
\includegraphics[width=2.5in]{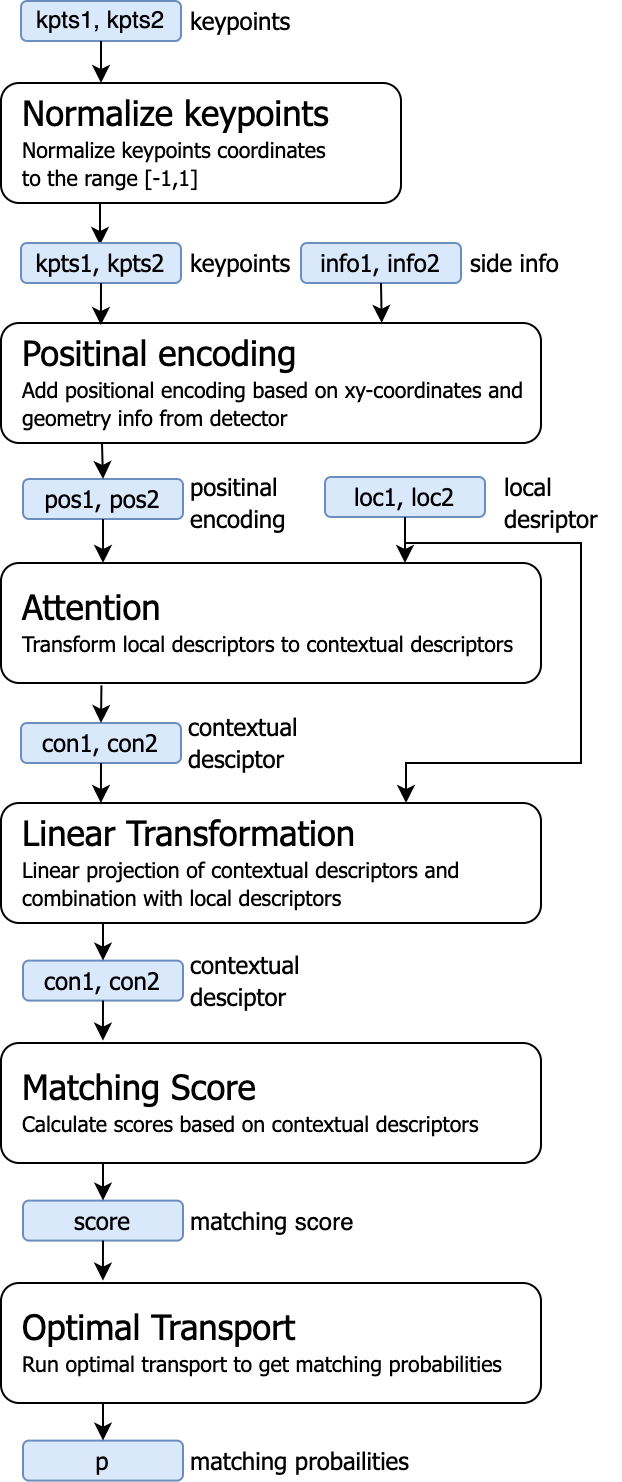}
\caption{\textbf{Matcher component of OpenGlue. Step-by-step algorithm}}
\vspace{-.5em}
\label{fig:stepbystep}
\end{figure}

 
We outline OpenGlue matcher part of our pipeline in Figure \ref{fig:stepbystep}. Each step in our interface is implemented separately, so it can be customized by user.

\subsection{Frameworks} 
We also utilize Kornia \cite{eriba2019kornia} an open-source differentiable computer vision library for PyTorch. 
Our implementation is based on PyTorch \cite{NEURIPS2019_9015}, an open-source library widely used in current industries and in the research community as it is flexible, well supported, easy to learn and enables the use of many complex features. PyTorch Lightning is a lightweight PyTorch wrapper for high-performance deep learning research that provides a full control and flexibility over the code while simplifying abstract standard operations. The code written with Lightning is clearer to read as many common functions are standardized. It is also robust and hardware agnostic. Moreover, in PyTorch Lightning, all code is self-contained enabling simple reproducibility. 
\par
We also utilize Kornia \cite{eriba2019kornia} an open-source differentiable computer vision library for PyTorch. This library is composed by a subset of packages containing operators that can be inserted within neural networks to train models to perform image transformations, epipolar geometry, depth estimation, and low-level image processing. The Kornia components that we use include kornia.feature (a module to perform feature detection and description), kornia.utils (image to tensor utilities and metrics), kornia.geometry (a geometric computer vision library to perform image transformations, 3D linear algebra and conversions using different camera models).

\subsection{Cashed features}\label{section:cached-feat}

In our framework, we introduce the ability to generate local features on the fly or cashing them in advance. The training speeds up immensely if the features are precomputed and saved, although this approach is expensive in terms of data storage. We suggest precomputing local features for all cases: SIFT, Dog-AffNet-HardNet and SuperPoint. Each of these methods has specific steps for features extraction.
\par
\textbf{SIFT:}
We use OpenCV version of SIFT to detect all possible keypoints.
We run non-maximum suppression when applicable, over the keypoints and then cache at most 2048 candidates with highest responses. During training at each iteration, we use a random cropping procedure for each input image. Not all 2048 candidates remain in the cropped region. After the crop, only 1024 keypoints are selected. 1024 candidates are selected at random for training and with highest responses for validation. This provides a slight augmentation at the training time.  
\par
\textbf{DoG-AffNet-HardNet:}
Again, we use OpenCV version of SIFT to make the keypoint detection. We apply the same NMS and keypoints selection strategies as with SIFT features. The only difference is that we use Kornia's implementation of AffNet \cite{AffNet2017} and HardNet \cite{HardNet2017} to estimate the local affine geometry of each keypoint and provide description.
\par
\textbf{SuperPoint:}
For SuperPoint we detect and describe with a pretrained model, run non-maximum suppression and choose 2048 candidates with highest response for cashing. The cropping procedure is the same as described above.
\par
During training, the number of keypoints per image decreases according to our cropping augmentation, followed by top-responses filtering, which propagates not more than 1024 keypoints per image. Since a batch can contain images with different number of detected keypoints, we stack them by selecting the minimum number of keypoints from a batch filtering out those with the lowest score.

\subsection{Inference}\label{section:inference}
At inference steps we skip the cropping step from data preprocessing and resize an image to the fixed size, $960~\times~720$. The number of maximum keypoints to be detected is configurable and can differ from the training setting for this parameter.

%% file: sections/Datasets.tex
\section{Datasets}

We used the MegaDepth \cite{MDLi18} dataset for training and validation, which is also used for training in SuperGlue. The dataset is split into images of separate scenes from different viewpoints. For each image, camera intrinsic and extrinsic parameters relative to the world coordinates are provided. Also, dense depth maps for the images are available, even though depth information is present only for part of the image. 
\par
For train set we use all the scenes that original SuperGlue selected for training and half the scenes, that they used for validation. The other half of SuperGlue's validation split we use for validating OpenGlue. 

\subsection{Preprocessing for training data}
During training we apply aspect-ratio preserving resize to the image and then crop it such that it has a final size fixed to M by N pixels. For images from training subset we apply random crop, while for validation we use center crop to ensure consistent validation results between training epochs. In all our experiments M is equal to 960 and N is 720.

\subsection{Evaluation data}
For evaluation we report the results on Phototourism test set from Image Matching Challenge \cite{Jin2020}, that was originally created to set a large-scale benchmark for for image matching. Phototourism is one of the three datasets from a challenge and it was included in a benchmark evaluation each year. 
\par
Phototourism data was obtained at different times, from different viewpoints, and with occlusions. It contains 26 photo-tourism image collections of popular landmarks originally collected by the Yahoo Flickr Creative Commons 100M (YFCC) \cite{YFCC2016} dataset and Reconstructing the world in six days. They range from ~100s to ~1000s of images. The training dataset has 15 scenes with ground-truth correspondences, while 3 of them (Reichstag, Sacre Coeur and Saint Peter's Square) are reserved for validation, and 12 separate scenes for testing.

%% file: sections/Experiments.tex
\section{Evaluation methods}\label{section:evaluation}

Establishing ground truth point-to-point correspondences requires accurate depth information. In MegaDepth case, we have noisy depth maps with many missing locations. The resulting set of ground truth correspondences consists of many ambiguous matches, which don't cause any problems during the training since we can ignore those matches when computing the loss. On the other hand, ignoring those matches at evaluation time can dramatically influence the metrics and result in a very inaccurate estimate of model performance. 
\par
A common strategy when evaluating the image matching method is to measure
its performance on the downstream task. In our case, we chose the task of
relative pose estimation since we can obtain accurate ground truth poses information for any image pair. Following the SuperGlue paper, we use the AUC of the pose error at (5\degree, 10\degree, 20\degree ) thresholds. We predict relative pose information by estimating the Essential matrix using methods based on RANSAC \cite{DBLP:journals/cacm/FischlerB81}. 
In our experiments we use MAGSAC++ \cite{barath2019magsac}. It is an estimator that uses iterative re-weighted least squares optimization approach combined with a soft threshold to make use of roughly correct, but imprecise correspondences. The pose error is computed by accounting for angular error in rotation and translation and taking the maximum of two. To analyze the performance of SuperGlue at the level of individual keypoints matches on MegaDepth validation set, we use precision (P) and matching score (MS) metrics.

\[ Precision (P) = \tfrac{True \ Positives}{ Predicted \ Matches} \]
\[ Matching Score (MS) = \tfrac{True \ Positives}{Total \ Keypoints} \]

The match is considered true positive if the corresponding epipolar error in a calibrated ray space is smaller than the predefined threshold 5e-4. This constant is chosen the same as in the original paper for the consistency purposes. We do not estimate the recall of matching on MegaDepth data since the depth information is not fully available for each image. Nevertheless, MS is used to estimate the method’s performance in terms of false negative predictions.
\par 

Image Matching Challenge has a slightly different approach to measuring the quality of the estimated poses. We use their metrics for testing on Phototourism, namely mean Average Accuracy(mAA) of camera pose error at 10\degree.

\section{Experiments}

We experiment with each custom component separately to show the impact for each case.
\par

\begin{table}[ht]
\footnotesize
	\caption{mAA@$10^\circ$ results on IMC2021 test set, mean average accuracy, Phototourism. First block: OpenGlue matching, second row block: features with SMNN matching, third block: SuperGlue and IMC2021 winners with submission ids.}
	\label{table:local_feat_phototest}
	\centering
		\begin{tabular}{clcc}
			& Method & 2K & 8K\\
			\toprule
			\multirow{10}{*}{\rotatebox[origin=c]{90}{Stereo}}
			& RootSIFT  + \textbf{OpenGlue} &  0.4160 & 0.4927\\
			& DoG-AffNet-HardNet (NMS) + \textbf{OpenGlue} & \textbf{0.4294} &  \textbf{0.5108}\\
			\cmidrule(l){2-4}\\
			& SuperPoint & 0.2964 & n/a \\
			& RootSIFT & 0.3149 & 0.4930 \\
			& DoG + HardNet & 0.3857 & 0.5543 \\
			& UprightRootSIFT & 0.3954 & 0.5100 \\
			& Upright DoG + HardNet & \textbf{0.4609} & \textbf{0.5727} \\
			\cmidrule(l){2-4}\\
			& LoFTR (\href{https://www.cs.ubc.ca/research/image-matching-challenge/2021/submissions/29f8cfce05f7db9c-loftr_v4/}{\#29f8cfce} 2021) & n/a & 0.6090\\
			& SuperPoint  + SuperGlue (\href{https://www.cs.ubc.ca/research/image-matching-challenge/2020/submissions/sid-00589-sp-k2048-nms3-refine2-r1600forcecubic-down128_sg-t.2-it150_degensac-th1.2/}{\#00589} 2020) & \textbf{0.5523} & n/a \\
			& (SP and DISK) + SuperGlue(\href{https://www.cs.ubc.ca/research/image-matching-challenge/2021/submissions/e90436771b444793-613json/}{\#e9043677} 2021) & n/a & \textbf{0.6397}\\
			\midrule[\heavyrulewidth]
			\multirow{10}{*}{\rotatebox[origin=c]{90}{Multiview}} 
				& RootSIFT  + \textbf{OpenGlue} &  0.6673 & 0.7118\\
			& DoG-AffNet-HardNet (NMS) + \textbf{OpenGlue} &  \textbf{0.6784} & \textbf{0.7155}\\
		
			\cmidrule(l){2-4}	\\
			& SuperPoint & 0.5464 & n/a \\
			& RootSIFT & 0.4682 & 0.6506 \\
			& UprightRootSIFT & 0.5623 & 0.6885 \\
			
			& DoG + HardNet & 0.5661 & 0.6090 \\
			& Upright DoG + HardNet & \textbf{0.6354} & \textbf{0.7230} \\
			
			\cmidrule(l){2-4}\\
			& LoFTR(\href{https://www.cs.ubc.ca/research/image-matching-challenge/2021/submissions/29f8cfce05f7db9c-loftr_v4/}{\#29f8cfce} 2021) & n/a & 0.7610\\
			& SuperPoint  + SuperGlue (\href{https://www.cs.ubc.ca/research/image-matching-challenge/2020/submissions/sid-00589-sp-k2048-nms3-refine2-r1600forcecubic-down128_sg-t.2-it150_degensac-th1.2/}{\#00589} 2020) & \textbf{0.7521} & n/a \\
		
			& (SP and DISK) + SuperGlue(\href{https://www.cs.ubc.ca/research/image-matching-challenge/2021/submissions/e90436771b444793-613json/}{\#e9043677} 2021) & n/a & \textbf{0.7857}\\
			\bottomrule
		\end{tabular}
\end{table}

\begin{figure*}
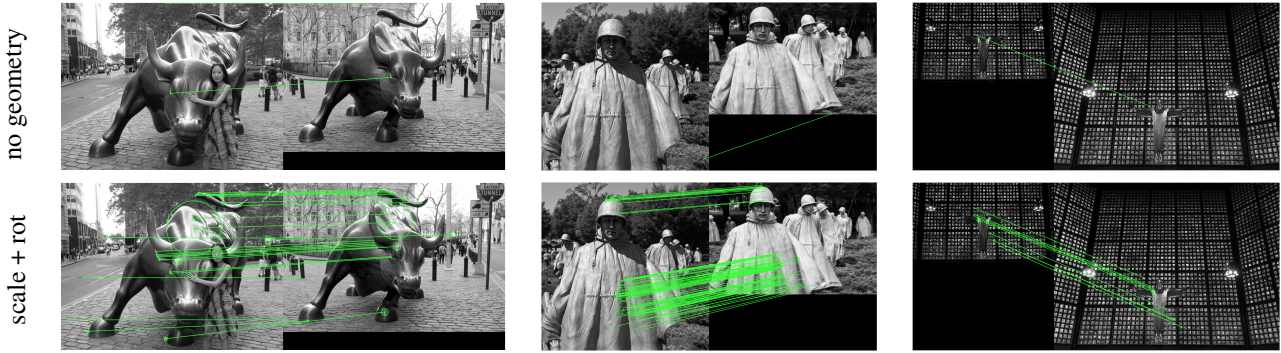

\centering
\begin{tabular}{cccc}
\rotatebox[origin=c]{90}{no geometry} 
    & \animage 
    & \animageone
    & \animagetwo  \\
\rotatebox[origin=c]{90}{scale + rot} 
    & \animagegeo
    & \animageonegeo
    & \animagetwogeo \\
\end{tabular}
\caption{Correct matches visualization on MegaDepth validation scenes for OpenGlue versions: without geometry vs. with scale and rotation. The shown cases significantly differ between the two models in the number of correctly found matches.}
\label{figure:correctmatches}
\end{figure*}

\begin{table}[!t]
\centering
\footnotesize
\caption{MegaDepth validation set. OpenGlue matcher with selected local features. We selected three types of approaches to test: entirely hand-crafted, a hand-crafted detector with a learned descriptor, and end-to-end learnable. Each type is separated in this table by a horizontal line. Bold text shows the best results among all configurations per each method type.}
\vspace{0.5em}

\begin{tabular}{lP{1.cm}P{1.cm}P{1.cm}}
\Xhline{2\arrayrulewidth}
Local Features & Matching Score & Precision & AUC 10\degree  \\
\Xhline{2\arrayrulewidth}
SIFT \& scale+rot & 0.1379 & 0.8644  & 0.3100\\
SIFT (NMS) \& scale+rot & \textbf{0.2154} & \textbf{0.9410} & \textbf{0.4167} \\
\hline
\\
DoG-AffNet-HardNet (NMS) & \textbf{0.2411} & \textbf{0.9426} & \textbf{0.4307} \\
\hline
\\
SuperPoint (KITTI) & 0.2112 & 0.9448 & 0.4004 \\
SuperPoint (COCO) & 0.1967 & 0.9086 & 0.3656 \\
SuperPoint (MagicLeap)  & \textbf{0.2489} & \textbf{0.9463} & \textbf{0.4491}\\
\hline
\end{tabular}
\label{table:megadepth_localfeat}
\end{table}

\textbf{Local Features:} 
We test OpenGlue in combination with various local features and verify the importance of the proposed improvements. We report the results on the  PhotoTourism test set, ref. Table \ref{table:local_feat_phototest} and on the MegaDepth validation subset, ref. Table \ref{table:megadepth_localfeat}. 
\par 
With our proposed changes, namely adding non-maximum suppression, OpenGlue with SIFT is capable to achieve a significant boost in results compared to the set-up without NMS. If geometry information about keypoints is included, tt shows even better performance than SuperPoint with open-source weights on MegaDepth validation. DoG-AffNet-HardNet provides better metrics than all SIFT experiments and is competitive against SuperPoint, showing higher scores for all cases except for MagicLeap SuperPoint weights. The best performance is achieved with the latter, however, the weights to this model are licensed and can only be used for academic purposes with full rights reserved by the previous authors. 
The lower performance of OpenGlue with SuperPoint (KITTI) and SuperPoint (COCO) can be due to not ideal reimplemenation of the open sourced model compared to the closed source original model. 
\par
The results on PhotoTourism are shown in Table \ref{table:local_feat_phototest}. We compare OpenGlue against Mutual Nearest Neighbors matcher with second nearest check (SMNN\cite{Jin2020}), SuperGlue\cite{sarlin20superglue} and LoFTR\cite{sun2021loftr}. OpenGlue shows an improvement over SMNN matches for SIFT and DoG-AffNet-Hardnet features when the number of detected keypoints is limited to 2048. The performance improvement vanishes when increasing number of keypoints to 8k. Furthermore, SIFT features combined with OpenGlue perform worse compared to SuperPoint + SuperGlue, which cat be partly explained by the two times smaller size of SIFT descriptors compared to SuperPoint. Yet, one advantage of OpenGlue combined with SIFT descriptors is that OpenGlue matcher is 4 times faster and required 4 times less memory compared to SuperGlue operating on SuperPoint features.


\par
\textbf{Attention:} We compare the performance of OpenGlue implemented with different attention mechanisms, namely softmax, linear and FAVOR+ approximation of softmax adapted from Performer model. As a feature descriptor we use SuperPoint trained on the KITTI dataset in all three scenarios. The performance results with metrics shown in Table \ref{table:attention}. 
\par 
The performance of OpenGlue significantly degrades, when changing attention from softmax to its efficient counterparts. Nevertheless, using linear or FAVOR+ attention allows utilizing much less GPU memory, which can be used to train the model with larger batch size. Also, FAVOR+ does not offer any additional advantage over standard linear attention, even though it requires a bit more memory and computations. 

\begin{table}[!t]
\centering
\footnotesize
\caption{Comparison of attention mechanisms in SuperGlue. MS - matching score. Last column represents relative amount of memory needed by OpenGlue for making inference on one pair.}
\vspace{0.5em}
\begin{tabular}{lP{0.7cm}P{1.0cm}P{1.5cm}P{1.5cm}}
\Xhline{2\arrayrulewidth}
Attention & MS & Precision & AUC 10\degree & Memory (\%) \\
\Xhline{2\arrayrulewidth}
Linear & 0.1580 & 0.8419 & 0.2930  & \textbf{66.12}\\
FAVOR+ & 0.1498 & 0.7029 & 0.2892  & 70.98\\
Softmax & \textbf{0.2112} & \textbf{0.9448} & \textbf{0.4004} & 100.00\\
\hline
\end{tabular}
\label{table:attention}
\end{table}

\par
\textbf{Using Keypoints Geometry in Positional encoding:} Three scenarios with SIFT with non-maximum suppression for keypoints are tested: without any side information about transformations, with scale encoded in positional encoding, with a combination of scale and rotation both included. The impact is shown in Table \ref{table:geometry}. The biggest boost comes from using the feature orientation, whereas scale does not impact results much. Nevertheless, combined with orientation, scale improves results noticeably. Therefore, we suggest including both of them. In Figure \ref{figure:correctmatches}, we visualize samples of correct matching predictions for the model with encoded geometry and the model without it. The improvement is visible by the increased number of correctly found matches when geometry was included.

\begin{table}[!t]
\centering
\footnotesize
\caption{Comparison of geometry information's impact for positional encoding on the MegaDepth validation set. SIFT is used for local feature detection. MS - matching score}
\vspace{0.5em}
\begin{tabular}{lP{0.8cm}P{1.0cm}P{0.8cm}P{0.8cm}P{0.8cm}}
\Xhline{2\arrayrulewidth}
Info type & MS & Precision & AUC 5\degree & AUC 10\degree & AUC 20\degree \\
\Xhline{2\arrayrulewidth}
none & 0.1819 & 0.8859 & 0.2420 & 0.3587 & 0.4806\\
scale & 0.1805 & 0.8932 & 0.2482 & 0.3704 & 0.4990 \\
rotation & 0.2089 & 0.9345 & 0.2773 & 0.4063 & 0.5371\\
scale + rotation & \textbf{0.2154} & \textbf{0.9410} & \textbf{0.2850} & \textbf{0.4167}& \textbf{0.5496}\\
\hline
\end{tabular}
\label{table:geometry}
\end{table}

\par
\textbf{Residual:} We launch two experiments to compare the effect of adding residual. For local feature detection and description we use SuperPoint with open-source weights trained on KITTI data with 256 descriptor dimensionality, other components include 9 stages for graph neural net with softmax attention. Results show a 2.16 \% increase in Matching score, refer to Table \ref{table:residual}. In addition to that, residual connection allows model to converge faster.

\begin{table}[t!]
\centering
\footnotesize
\caption{Comparison of OpenGlue trained with and without residual on MegaDepth validation. Local features extracted with SuperPoint pre-trained on KITTI. MS - matching score}
\vspace{0.5em}
\begin{tabular}{lP{0.8cm}P{1.0cm}P{0.8cm}P{0.8cm}P{0.8cm}}
\Xhline{2\arrayrulewidth}
 & MS & Precision & AUC 5\degree & AUC 10\degree & AUC 20\degree \\
\Xhline{2\arrayrulewidth}
No residual & 0.1896 & 0.9398 & 0.2650 & 0.3878 & 0.5137\\ 
With Residual & \textbf{0.2112} & \textbf{0.9448} & \textbf{0.2747} & \textbf{0.4004} & \textbf{0.5270} \\
\hline
\end{tabular}
\label{table:residual}
\end{table}

%% file: sections/Conclusions.tex
\section{Conclusion}

OpenGlue is an end-to-end open-source learnable framework developed for various industrial and academic applications. We built it upon a SuperGlue architecture and customized its main components. In this work, we observed that SuperGlue effect significantly depends on the type of local features used. OpenGlue increases the matching performance of non-learnable (SIFT) and partially learnable (DoG-AffNet-HardNet) local features. 
To achieve this metric boost, first, we implemented non-maximum suppression over the keypoints for each detector. Secondly, we included geometry information about keypoints, such as scale and rotation, into positional encoding in training phase. Finally, adding a mechanism for combination of local descriptor with context-aware descriptor gave slight improvements.
\par
We show a trade-of between memory consumption and the performance when using various optimized attention types. Linear and FAVOR+ attention mechanisms speed up the performance immensely, however, produce significantly worse results.